\documentclass{article} %
\usepackage{iclr2026_conference,times}
\usepackage{xcolor}

\usepackage{float}
\usepackage{booktabs}
\usepackage{wrapfig}
\usepackage{capt-of}
\usepackage[most]{tcolorbox}
\usepackage[table]{xcolor}
\definecolor{sigpos}{RGB}{200,235,200}
\definecolor{signeg}{RGB}{245,200,200}
\definecolor{nonsig}{RGB}{225,225,225}
\newcommand{\dpos}[1]{\cellcolor{sigpos}$\mathbf{#1}$}
\newcommand{\dneg}[1]{\cellcolor{signeg}$\mathbf{#1}$}
\newcommand{\dns}[1]{\cellcolor{nonsig}$#1$}

\definecolor{tealframe}{HTML}{5BA4A4}
\definecolor{tealback}{HTML}{F5FAFA}

\newtcolorbox{takeaways}[1][Takeaways]{
  enhanced,
  colback=tealback,
  colframe=tealframe,
  coltitle=white,
  fonttitle=\bfseries\large,
  title=#1,
  arc=2pt,
  boxrule=1pt,
  left=6pt, right=6pt, top=6pt, bottom=6pt,
}

\usepackage{amsmath,amsfonts,bm}

\def\eqref#1{equation~\ref{#1}}

\def\1{\bm{1}}

\DeclareMathAlphabet{\mathsfit}{\encodingdefault}{\sfdefault}{m}{sl}
\SetMathAlphabet{\mathsfit}{bold}{\encodingdefault}{\sfdefault}{bx}{n}

\usepackage{hyperref}
\definecolor{linkblue}{HTML}{356792} %
\hypersetup{
  colorlinks=true,
  linkcolor=linkblue,   %
  citecolor=linkblue,   %
  urlcolor=linkblue,    %
}
\usepackage{url}
\usepackage{graphicx}

\usepackage{booktabs}

\title{{\Large Where and Why is Grounded Reasoning Useful?}}

\author{%
David Yan$^{1,2}$,
Le Xue$^{1}$,
Forrest Huang$^{1}$,
Qiuyi Zhang$^{1}$, \\
\bf Dustin Tran$^{1}$,
Andrew M.~Dai$^{1}$,
Yinfei Yang$^{1}$ \\\\
$^{1}$ Elorian \quad
$^{2}$ Princeton University
}

\definecolor{RoyalBlue}{RGB}{65,105,225}

\iclrfinalcopy %

\begin{document}

\maketitle

\begin{abstract}

 Grounded reasoning, where a model outputs text coordinates of relevant locations in chain-of-thought reasoning, is a common strategy for solving complex visual reasoning problems. However, the benefits and limitations of grounded reasoning are understudied. We study where and why grounded reasoning is most useful. To do so, we introduce a chain-of-thought analysis method to measure how much a model uses its outputted grounding in its thinking trace and perform a detailed ablation study of a grounded reasoning training recipe. We find that grounded reasoning improves counting but shows no significant gain over baselines without coordinate grounding across a wide range of spatial reasoning and perception benchmarks. Using several synthetic tasks, we show that grounded reasoning is most useful when a task requires undescribable references, which are visual concepts that are difficult to describe in language. However, we observe that many existing visual reasoning benchmarks are solvable with only describable text references. To close this gap, we introduce BrickVision, a new benchmark that requires undescribable visual references to solve. 
 
 \vspace{0.5em}
\noindent\textbf{Project Page:} \url{https://elorian.ai/blog/why-grounding}

\end{abstract}

\begin{figure}[h]
    \centering
    \includegraphics[width=1\linewidth]{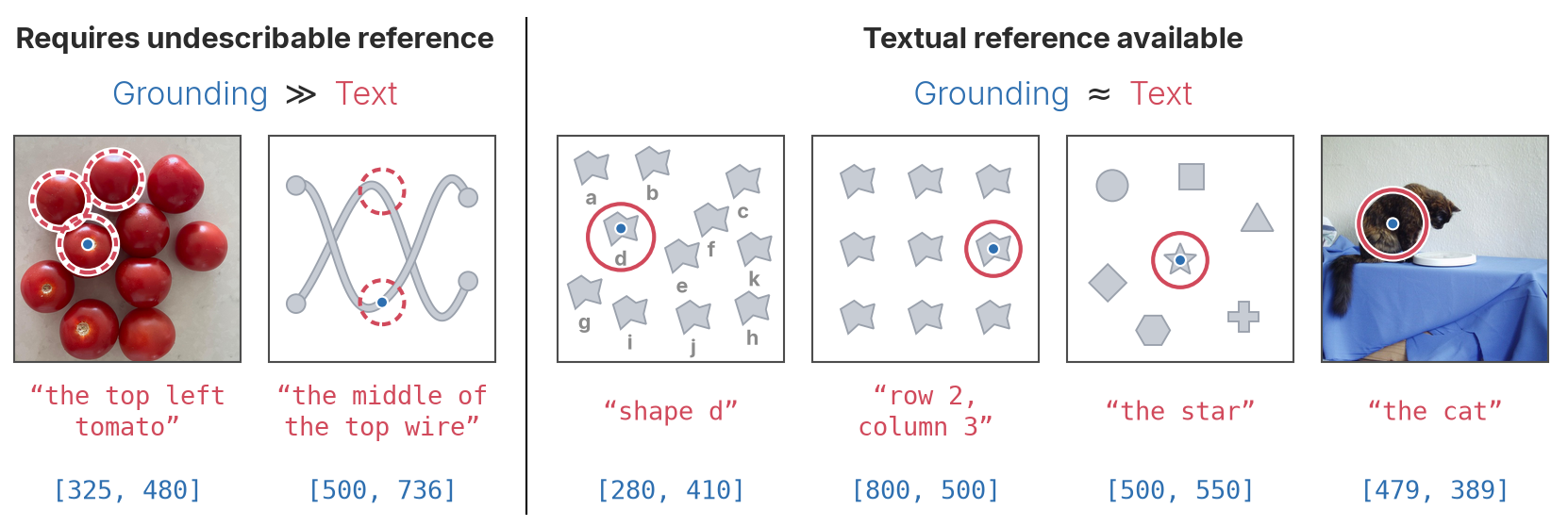}
    \caption{\textbf{Visual grounding is useful as an undescribable reference}. When objects in a task are uniquely identifiable in text (e.g. through a label, descriptor, or position), we say that a \textit{textual reference} is available. When textual references are available, grounded reasoning performs no better than text reasoning. When no unique text description is available (e.g. in dense counting or tracing tasks), we say the task requires maintaining \textit{undescribable references}. Grounded reasoning significantly outperforms text reasoning when the task requires undescribable references.}
    \label{fig:teaser}
\end{figure}
\section{Introduction} While solving visual tasks, humans shift their gaze between objects pertaining to the problem. Grounded reasoning models mimic this behavior by enabling a vision language model (VLM) to output text coordinates while reasoning. Grounded reasoning has a number of appealing properties. It can reduce hallucinations by anchoring the VLM to certain regions of the image and serves as an interpretable way of diagnosing model failures. Moreover, recent text-coordinate-based grounded reasoning methods do not require reprocessing of cropped images or replaying of visual features\footnote{Throughout this paper, we use ``grounded reasoning''  to refer to models and techniques that only output text coordinates during reasoning, as opposed to adding cropped images or features into context.}. For these reasons, training recipes for grounded reasoning have received broad interest \citep{lu2026think, wang2026traceableevidenceenhancedvisual,  zhang2026thinkingvisualgrounding, sarch2026groundedreinforcementlearningvisual}.

However, it remains unclear where and why grounded reasoning is most useful. For instance, recent results from iVGR \citep{zhang2026ivgrinternalizingvisuallygrounded} suggest that, on certain tasks, text-only reasoning modes can outperform grounded reasoning. On the other hand, works such as Thinking with Visual Primitives \citep{lu2026think} find improvements across a range of general VQA tasks. As such, an important goal is to understand which tasks necessitate the additional complexity of grounded reasoning and which domains can be solved with natural language reasoning. In this work, we investigate the following question: \textit{where and why is grounded reasoning actually necessary to solve a task}?

\begin{figure}[h!]
    \centering
    \includegraphics[width=1.0\linewidth]{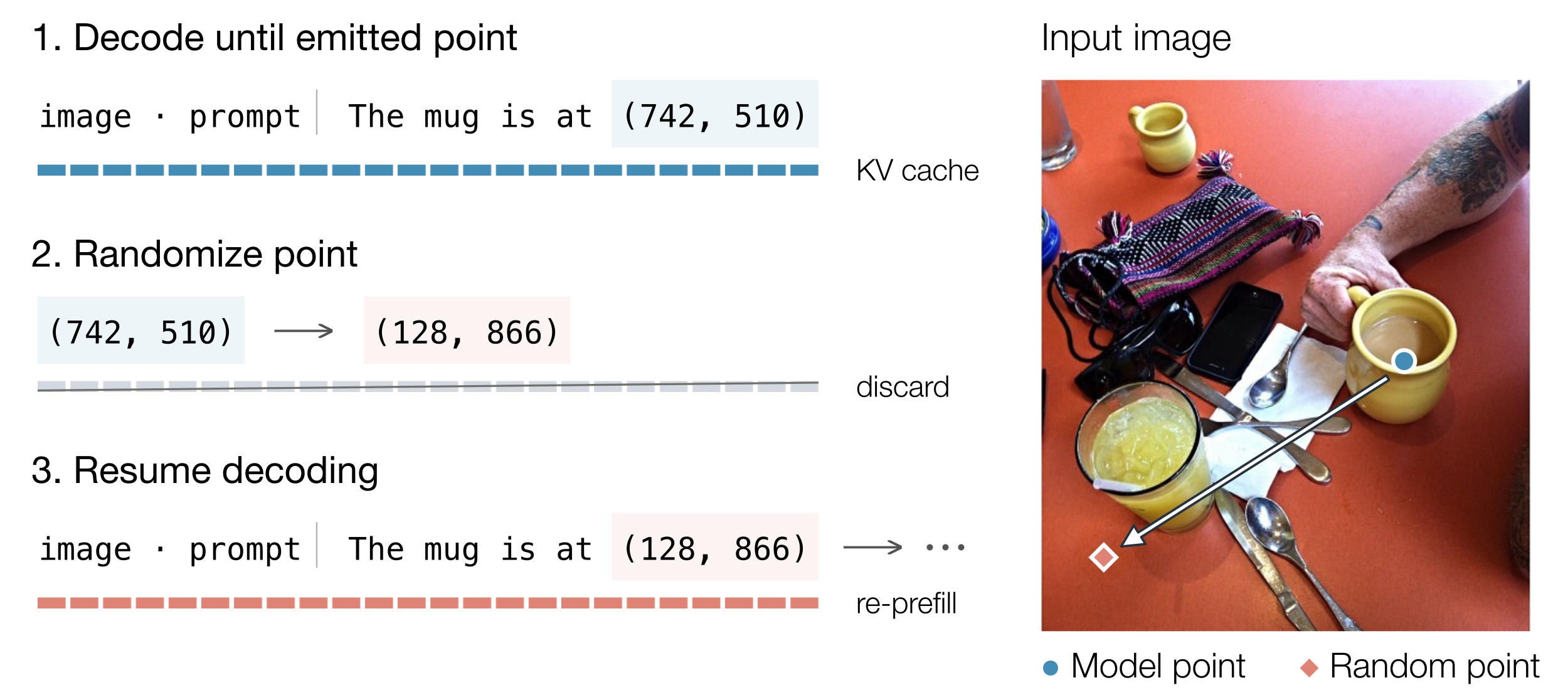}
    \caption{\textbf{Decode Grounding Corruption}. We introduce a CoT analysis technique to analyze whether a model uses its grounding. Instead of editing CoT after generation, we randomize the model's outputted points during decoding and reset the KV-cache.}
    \label{fig:decode_rand}
\end{figure}

We first perform a training-free study of whether existing models actually rely on their intermediate grounding outputs to solve tasks. Existing chain-of-thought (CoT) interpretability techniques edit or insert mistakes into the chain-of-thought and examine if the model answer changes \citep{lanham2023measuringfaithfulnesschainofthoughtreasoning}. We adapt this approach to measure grounding reliance, as opposed to general CoT faithfulness. Simply removing or randomizing points after the CoT is generated is an insufficient test, because the model might have derived information from the coordinates in subsequent text reasoning. We therefore propose a test for measuring grounding usage that repeatedly randomizes the CoT's emitted points during inference (Figure~\ref{fig:decode_rand}). Our method shows that many existing open-source grounded reasoning models do not rely on grounding for perception and spatial reasoning tasks, with performance remaining generally unchanged after coordinate randomization (Table~\ref{tab:decode-corr}).

In order to understand this phenomenon, we train our own grounded reasoning models and carefully study the effect of different training recipes on downstream performance. We find that the usage of model grounding is sensitive to the accuracy of the grounding coordinates during SFT, but is also generally uncorrelated with spatial reasoning  performance. Moreover, even with high-quality reasoning traces and grounding annotations, we find that grounded reasoning is not helpful when compared to a strong text-only reasoning baseline on perception and reasoning benchmarks (Table~\ref{tab:sft-rl}). In contrast, we find that grounded reasoning is helpful for counting benchmarks. How can we explain why grounding is helpful for counting tasks but not for spatial reasoning or perception?

We hypothesize that the major value of grounded reasoning is when a task requires maintaining \textit{undescribable references}, which are visual concepts that are difficult to describe in language (Figure~\ref{fig:teaser}). To demonstrate this, we develop a suite of evaluations across tracing, counting, and tracking tasks. For each benchmark, we train a model on  high-quality, in-domain chain of thought traces for both text-only and grounded reasoning. When visual concepts are hard to describe with text, the benefit of grounding is large. Conversely, when labels are added to the task, the marginal value of grounding over pure-text reasoning drops significantly (Figure~\ref{fig:probe}).

We observe that most existing visual reasoning benchmarks struggle to demonstrate the benefit of grounded reasoning because they are mostly solvable with describable references. To close this gap, we use insights from our study to design BrickVision, a challenging new visual reasoning benchmark that requires maintaining undescribable references. On BrickVision, models such as GPT-5.6 Sol and Gemini 3.8 Flash score just 18.9\% and 15.5\%, respectively (Table~\ref{tab:lego-base}). GPT-6 Astra is significantly stronger than all other existing models, with a score of 68.7\%. Our main contributions are as follows:

\begin{enumerate}
    \item We introduce a technique to measure grounding usage in existing reasoning models and find that many open models do not use their generated coordinates when reasoning.

    \item We perform a detailed ablation study of the grounding training recipe across spatial reasoning, fine-grained perception, and counting tasks. We find that grounded reasoning outperforms natural language reasoning on tasks that require visual references difficult to describe in language, such as counting and tracking.
    
    \item We introduce BrickVision, a challenging benchmark designed to measure a model's capability to reason with undescribable visual references. Many frontier models perform poorly on BrickVision, scoring under 30\%.

\end{enumerate}

\section{Related Work}

\textbf{Grounded reasoning models.} A longstanding issue for VLMs has been over-reliance on textual priors \citep{asadi2026mirageillusionvisualunderstanding, chen2024rightwayevaluatinglarge, vo2026visionlanguagemodelsbiased, bai2025hallucinationmultimodallargelanguage, kancheti2026chainofthoughtdegradesvisualspatial}, which weakens perception and visual reasoning performance. As such, a large variety of techniques exist to enable models to reground to an image during reasoning. One common approach is to format the re-grounding operation as a tool call, which then runs a crop or zoom operation \citep{zheng2026deepeyesincentivizingthinkingimages}. However, repeatedly processing images through tool calls adds significant overhead. An alternative approach is to replay a subset of the original image's features into the context \citep{wang2026vgrvisualgroundedreasoning}. This method requires specialized training and modules to process image tokens, adding significant complexity to model design. Coordinate-only grounding approaches, where a model simply interleaves text coordinates during reasoning, enable a model to reason over grounding without accumulating image tokens or requiring any bespoke architectural changes. These properties are highly appealing, and as such, a number of recent works have focused on developing training recipes for grounded reasoning \citep{lu2026think, wang2026traceableevidenceenhancedvisual, zhang2026ivgrinternalizingvisuallygrounded, zhang2026thinkingvisualgrounding, perceptron2026visualthinking, sarch2026groundedreinforcementlearningvisual}. For these reasons, we opt to primarily study coordinate-only grounding over other grounded reasoning variants.

\textbf{Analysis of grounded reasoning}. Recent works have provided some insight into the strengths and weaknesses of grounding. iVGR \citep{zhang2026ivgrinternalizingvisuallygrounded} finds that when grounded reasoning models are prompted not to output coordinates, their performance generally increases across a large range of benchmarks. They attribute this phenomenon to the model making errors in grounding. We find contradictory results: model performance generally remains stable even when all grounding coordinates are forced to be incorrect by our decode randomization. \cite{perceptron2026visualthinking} performs a similar analysis of the grounded reasoning recipe but only considers the single synthetic task of Set. We benchmark grounding across a variety of real-world evaluation benchmarks and develop several synthetic tasks to illustrate the strength of grounding. \cite{lu2026think} claims that grounded reasoning is helpful for topological reasoning tasks like wire-tracing but does not compare against a matched text reasoning baselines. We provide controlled experiments against text reasoning and propose a more general explanation for where grounded reasoning is most useful.

\textbf{Visual reasoning benchmarks.} Precise perception and spatial understanding are widely observed weaknesses of VLMs. As such, a wide range of reasoning benchmarks seek to evaluate VLMs on such capabilities \citep{Liu_2025, lee2026multihopspatialmultihopcompositionalspatial, zhang2025mmerealworldmultimodalllmchallenge, du2024embspatialbenchbenchmarkingspatialunderstanding, jia2026omnispatialcomprehensivespatialreasoning}. 
 \cite{lu2026think}, \cite{chen2026babyvisionvisualreasoninglanguage}, and \cite{Berman_2025} also consider the tracing task. However, we use it as a controlled training environment to demonstrate the effectiveness of grounded reasoning over text reasoning and direct fine-tuning, as opposed to an evaluation of the zero-shot capabilities of VLMs.

 Most similar to our work is BabyVision \citep{chen2026babyvisionvisualreasoninglanguage}, which  also considers visual tasks that are difficult to describe in language. However, many BabyVision tasks accept text answer formats and therefore contain textualizable references for objects (e.g., which puzzle piece in the image, labeled A, B, C, D, would fit into the gap). In contrast, objects in BrickVision are unlabeled and placed in irregular locations; as such, undescribable references are necessitated. Moreover, BrickVision uses points as the answer format, which requires greater precision compared to a multiple-choice question and yields a more interpretable analysis of failure modes.

\section{Where Is Grounding Useful?}

\subsection{Do Models Use Their Grounding?}

\textbf{Decode-time Grounding Randomization}. We first aim to determine whether existing grounded reasoning models rely on their grounding during reasoning. One approach would be to have the model emit its full reasoning trace and answer, and then measure if the model changes its answer if the coordinates are randomized. However, this approach doesn't account for information that the model may have already derived from the grounding during further text reasoning. For example, a model that emits the coordinate of two objects may use those coordinates to reason ``object A is at [x1,y1] and object B is at [x2, y2], therefore A is to the left of B.'' Randomizing just the coordinates after the full reasoning trace is generated would still preserve ``therefore A is to the left of B,'' which the model could still use to solve the problem.

Therefore, to measure whether existing model checkpoints actually use their grounding coordinates, we introduce a test that corrupts coordinates \textit{during decoding}. Whenever a model generates a coordinate, we randomize the coordinate and reset the KV-cache using the edited prefix. As such, all future generation is conditioned on a randomized coordinate, which gives a more accurate signal on the effect of the coordinates during reasoning. This method is similar to the ``Adding Mistakes'' intervention in \cite{lanham2023measuringfaithfulnesschainofthoughtreasoning}, but is executed programmatically rather than by an LLM inserting a mistake. Moreover, we perform this intervention multiple times per completion, as opposed to doing a single intermediate mistake insertion. For each question, we sample 20 completions at temperature 1.0, and report the change in mean performance (\% acc) after randomization.

\begin{table}[h!]
\centering
\small
\begin{tabular}{lcccc}
\toprule
Model & MultihopSpatial & SpatialMQA & V$^\ast$ & CountQA \\
\midrule
\multicolumn{5}{l}{\textit{Open models}} \\
TreeVGR-7B-CI & $-0.62 \pm 0.33$ & $-0.50 \pm 0.28$ & $+0.55 \pm 0.71$ & $-0.18 \pm 0.26$ \\
TreeVGR-7B-RL & $-0.17 \pm 0.22$ & $-0.21 \pm 0.19$ & $-0.08 \pm 0.35$ & $-3.19 \pm 0.22$ \\
iVGR-8B      & $-0.93 \pm 0.29$ & $-0.41 \pm 0.32$ & $-0.99 \pm 0.51$ & $-2.64 \pm 0.29$ \\
DeepEyes-7B  & $+0.63 \pm 0.24$ & $+0.02 \pm 0.26$ & $-0.68 \pm 0.38$ & $-0.50 \pm 0.24$ \\
\midrule
\multicolumn{5}{l}{\textit{Ours} (format type)} \\
Appended     & $-1.64 \pm 0.31$ & $-0.61 \pm 0.35$ & $-2.36 \pm 0.59$ & $-0.23 \pm 0.29$ \\ 
\textbf{Reflection}   & $-3.63 \pm 0.30$ & $-2.50 \pm 0.35$ & $-4.16 \pm 0.62$ & $-3.88 \pm 0.28$ \\
\bottomrule
\end{tabular}
\caption{\textbf{Grounding usage experiments.} We report change in mean performance (\% acc) for our coordinate corruption experiments. Many open models don't rely on their grounding for spatial reasoning and perception. Our retrained models (below) are more reliant on their grounding outputs.}
\label{tab:decode-corr}
\end{table}

\textbf{Existing recipes don't necessarily lead to grounding usage.} We use our grounding randomization technique to analyze grounded reasoning models, such as TreeVGR, iVGR, and DeepEyes, across several benchmarks. We observe that, in general, open models do not significantly use their grounding (Table~\ref{tab:decode-corr}), except on counting tasks. This finding re-contextualizes results from iVGR, which finds that models prompted not to use grounding see performance increases. \cite{zhang2026ivgrinternalizingvisuallygrounded} posits that this difference can be attributed to models making grounding errors during reasoning, which the models then incorrectly reason over. However, our results demonstrate that models don't experience significant performance degradation even when all grounding is forced to be incorrect in our corruption experiments, suggesting that this explanation is incomplete.

\begin{figure}
    \centering
    \includegraphics[width=1.0\linewidth]{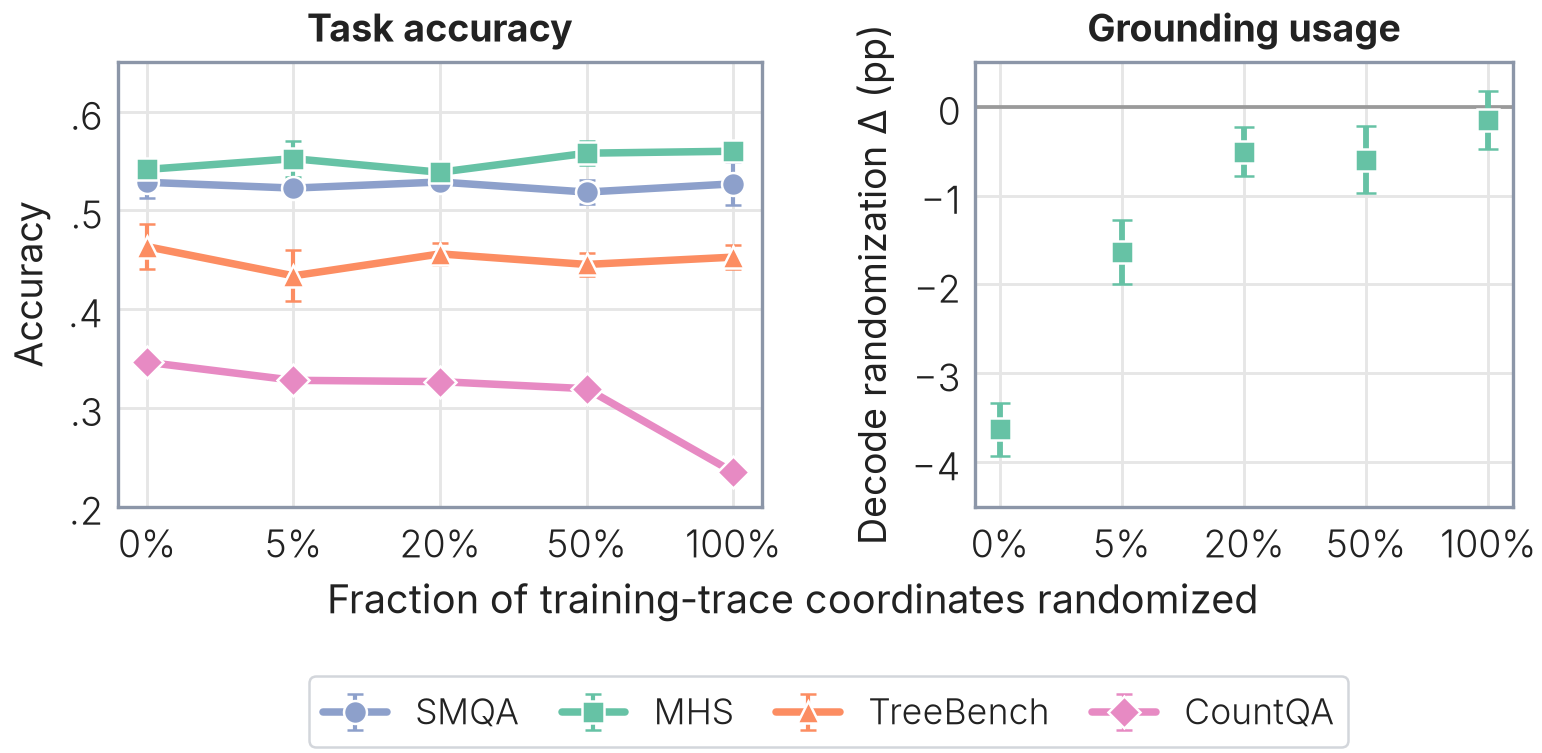}
    \caption{\textbf{Training Coordinate Randomization}. We randomize varying percentages of coordinates in the SFT set and report task performance and grounding usage. While grounding usage decreases with more randomization, task performance remains relatively flat, except on counting.}
    \label{fig:train_rand}
\end{figure}
\subsection{Studying the Grounded Reasoning Recipe}
\textbf{Problem setting and evaluation}. Our goal is to study various aspects of the training recipe for grounded reasoning. We train models across many variations of a training recipe and evaluate them over a wide range of public benchmarks where we expect grounding might help. We categorize each benchmark into one of the following broad task types: spatial reasoning, perception, and counting. For spatial reasoning, we evaluate on SpatialMQA (SMQA) \citep{Liu_2025}, MultihopSpatial (MHS) \citep{lee2026multihopspatialmultihopcompositionalspatial}, EmbSpatial \citep{du2024embspatialbenchbenchmarkingspatialunderstanding}, VSR \citep{liu2023visualspatialreasoning}, OmniSpatial \citep{jia2026omnispatialcomprehensivespatialreasoning}, and ERQA \citep{geminiroboticsteam2025geminiroboticsbringingai}. For perception, we evaluate on V* \citep{wu2023vguidedvisualsearch}, MME-RealWorld-Lite \citep{zhang2025mmerealworldmultimodalllmchallenge}, and TreeBench \citep{wang2026traceableevidenceenhancedvisual}. For counting, we evaluate on CountQA \citep{tamarapalli2025countqamllmscountwild} and FSC-147 \citep{ranjan2021learningcount}. We use a randomly selected subset of MultihopSpatial (1500 / 4500) for faster experiment iteration. We find that this sample size is sufficiently large to get reliable, comparative evaluations of models.

Unless otherwise specified, we train 3 checkpoints with random seeds. For each trained checkpoint, we evaluate 3 times with separate random seeds at a temperature of 1.0 and report the mean result for each benchmark. On the counting benchmarks (CountQA and FSC-147), we instead decode each checkpoint greedily, as we find that this improves performance significantly. To evaluate the differences between grounded and ungrounded checkpoints, we report the mean performance delta of the grounded performance minus the ungrounded performance (denoted $\bar{\Delta}$). We then compute a 95\% confidence interval for the delta (see Appendix~\ref{app:stats} for more detail). We consider a delta significant if the confidence interval does not include zero.

\textbf{Dataset construction}. We observe that existing open datasets have incorrect or low-quality grounding and/or reasoning traces. For instance, the TreeVGR SFT dataset has bounding boxes that are extremely large and generally inaccurate (Appendix~\ref{fig:vgrsample}). On the other hand, datasets from Thinking with Visual Grounding \citep{zhang2026thinkingvisualgrounding} have accurate grounding but excessively long thinking traces. As such, we construct our own high-quality SFT datasets based on \cite{zhang2026thinkingvisualgrounding}'s annotated SpatialMQA and MultihopSpatial sets. We first generate ungrounded thinking traces by prompting Gemini 3.5 Flash \citep{google2026gemini35flash} to produce a rationale for the solution without grounding. We then provide the QA, trace, and ground truth grounding annotations to Gemini 3.5 Flash and prompt the model to edit the trace to incorporate the grounding annotations. We choose to use point annotations over boxes in our grounded reasoning data, as we find points generally perform better than boxes across most benchmarks (Appendix~\ref{tab:sft-pointbox}). Additionally, we filter out samples that contain incorrect coordinates or large quantities of irrelevant grounding coordinates. This yields a high-quality grounded SFT dataset of 6,368 samples. For RL, we use the TreeVGR RL dataset, which is composed of questions from the $V^*$ train set \citep{wu2023vguidedvisualsearch} and VisDrone \citep{zhu2021detectiontrackingmeetdrones}. We additionally supplement the RL training set by rolling out both a text and a grounded reasoning SFT checkpoint on the SFT QA set and adding any question with $\text{pass@8} \in [0.2, 0.8]$ under both checkpoints.

\textbf{Training}. Following prior work \citep{wang2026traceableevidenceenhancedvisual, zhang2026ivgrinternalizingvisuallygrounded}, we first perform a cold-start SFT stage before reinforcement learning. Unless otherwise specified, we initialize from the Qwen3-VL 8B~\citep{bai2025qwen3vltechnicalreport} instruct model. More recent models, such as Qwen3.5~\citep{qwen3.5}, are explicitly trained for grounded reasoning. To isolate the effect of our training recipe, we use an older model. For RL, we apply a format, correctness, and grounding reward. We do not apply the grounding reward for text reasoning checkpoints.  We train using LlamaFactory \citep{zheng2024llamafactory} and veRL~\citep{sheng2024hybridflow}. Our RL training setup follows TreeVGR, but we use a slightly altered grounding reward from \cite{jiang2025detectpointprediction} (see Appendix~\ref{app:training}).

\begin{wraptable}{r}{0.5\textwidth}
\centering
\small
\setlength{\tabcolsep}{3pt}
\resizebox{\linewidth}{!}{%
\begin{tabular}{lc cc}
\toprule
Benchmark & Appended & Reflection & $\Delta$ \\
\midrule
SMQA  & 0.529 & 0.527 & \dns{-0.2 \pm 1.2} \\
MHS  & 0.544 & 0.539 & \dns{-0.5 \pm 1.2} \\
V$^*$  & 0.839 & 0.844 & \dns{+0.5 \pm 4.5} \\
MME-RW & 0.533 & 0.531 & \dns{-0.2 \pm 1.0} \\
TreeBench & 0.451 & 0.467 & \dns{+1.6 \pm 2.0} \\
CountQA  & 0.300 & 0.327 & \dpos{+2.7 \pm 1.9} \\
FSC-147 & 0.073 & 0.103 & \dpos{+3.0 \pm 1.6} \\
\bottomrule
\end{tabular}}
\caption{\textbf{Appended vs reflection performance}. The reflection reasoning style yields significantly better performance on counting.}
\label{tab:sft-paste}
\end{wraptable}
\subsubsection{Results}

\begin{table}[t]
\centering
\small
\setlength{\tabcolsep}{4pt}
\begin{tabular}{lc ccc ccc}
\toprule
 & & \multicolumn{3}{c}{SFT} & \multicolumn{3}{c}{RL} \\
\cmidrule(lr){3-5} \cmidrule(lr){6-8}
Benchmark & $n$ & Ungrounded & Grounded & $\Delta$ & Ungrounded & Grounded & $\Delta$ \\
\midrule
\multicolumn{8}{l}{\emph{Spatial reasoning}} \\
SMQA & 1076 & 0.537 & 0.527 & \dns{-1.0 \pm 2.1} & 0.571 & 0.562 & \dns{-0.9 \pm 1.8} \\
MHS & 1500 & 0.551 & 0.539 & \dns{-1.3 \pm 1.3} & 0.615 & 0.588 & \dneg{-2.7 \pm 1.9} \\
EmbSpatial & 3640 & 0.730 & 0.733 & \dns{+0.2 \pm 0.7} & 0.791 & 0.775 & \dneg{-1.7 \pm 0.9} \\
VSR & 1222 & 0.803 & 0.801 & \dns{-0.2 \pm 1.5} & 0.789 & 0.733 & \dneg{-5.5 \pm 1.3} \\
OmniSpatial & 1533 & 0.473 & 0.470 & \dns{-0.3 \pm 1.2} & 0.510 & 0.518 & \dns{+0.9 \pm 1.5} \\
ERQA & 400 & 0.393 & 0.396 & \dns{+0.3 \pm 2.7} & 0.432 & 0.423 & \dns{-0.9 \pm 3.0} \\
\midrule
\multicolumn{8}{l}{\emph{Fine-grained perception}} \\
V$^*$ & 191 & 0.834 & 0.844 & \dns{+1.0 \pm 2.7} & 0.876 & 0.898 & \dns{+2.3 \pm 2.7} \\
MME-RW & 1919 & 0.533 & 0.531 & \dns{-0.3 \pm 1.3} & 0.569 & 0.575 & \dns{+0.6 \pm 1.0} \\
TreeBench & 405 & 0.449 & 0.467 & \dns{+1.8 \pm 3.1} & 0.484 & 0.483 & \dns{-0.1 \pm 2.7} \\
\midrule
\multicolumn{8}{l}{\emph{Counting}} \\
CountQA & 1528 & 0.291 & 0.327 & \dpos{+3.6 \pm 2.4} & 0.292 & 0.326 & \dpos{+3.4 \pm 3.3} \\
FSC-147 & 1190 & 0.077 & 0.103 & \dpos{+2.6 \pm 1.3} & 0.067 & 0.088 & \dpos{+2.1 \pm 1.4} \\
\bottomrule
\end{tabular}
\caption{\textbf{Training recipe comparison}. We report the performance of grounded reasoning and text-only reasoning training recipes after SFT and RL training phases. Deltas that exclude 0 from the confidence interval are highlighted. We find that grounding provides a statistically significant benefit on counting, but not on spatial reasoning or perception.}
\label{tab:sft-rl}
\end{table}

\textbf{Coordinate accuracy affects grounding usage, but not accuracy}. We first study how the accuracy of the grounding annotations in training affects downstream performance and grounding usage. We randomize varying percentages of the coordinates in our SFT training set and examine the downstream SFT checkpoint performance on SpatialMQA, MultihopSpatial, TreeBench, and CountQA. We also perform the coordinate randomization test on each checkpoint on MultihopSpatial. We report our results in Figure \ref{fig:train_rand}. We find that grounding usage decreases significantly as a larger percentage of coordinates is randomized, with the delta going from $-3.63$ at $0\%$ randomized to $-0.15$ at $100\%$ randomized. This result suggests that noise in the training data may cause the model not to use its grounding. It also supports design decisions in \cite{lu2026think}, which note the importance of having high-quality ground truth in the SFT stage. Furthermore, while grounding usage decreases, we find that performance generally remains flat across spatial reasoning and fine-grained perception benchmarks, but decreases monotonically for CountQA. The lack of correlation between grounding usage and performance suggests that existing perception and spatial reasoning benchmarks do not benefit from better grounded reasoning. On the other hand, counting benefits significantly from effective grounded reasoning.

\textbf{Reasoning templates affect grounding usage}. We also ablate various strategies commonly used to insert grounding data in reasoning traces. We consider two approaches: \textit{Appended}, where grounding coordinates are appended when relevant objects are mentioned during reasoning, and \textit{Reflection}, where the traces contain a reflection phrase ``Let's look at [coordinate]'' (see Appendix~\ref{fig:format} for examples). We find that reflection results in significantly more grounding usage than appending grounding coordinates (Table~\ref{tab:decode-corr}) and also much better performance on counting (Table~\ref{tab:sft-paste}). These results show that ``hinting'' the model to use its grounding can be an effective strategy for improving performance. As such, we opt to use the reflection format in all other experiments.

\textbf{Grounding recipes are not generally better than text-only recipes}. We report the performance of grounded and ungrounded training recipes after SFT and after RL in Table~\ref{tab:sft-rl}. After SFT, the delta between grounded and ungrounded recipes is positive and significant for both counting benchmarks (+3.6 on CountQA and +2.6 on FSC-147), but is not significant on any spatial reasoning or fine-grained perception benchmarks. Similarly, after RL, only the counting benchmarks retain a significant positive delta, while we see regressions for MultihopSpatial, EmbSpatial, and VSR. These results suggest that even with high-quality training data and high grounding usage, grounded reasoning is generally helpful for counting tasks but not for spatial reasoning and perception. Based on these results, we seek to understand why grounded reasoning benefits some tasks, but not others.

\section{Why Is Grounding Useful?}
\label{sec:why}

\begin{figure}
    \centering
    \includegraphics[width=1\linewidth]{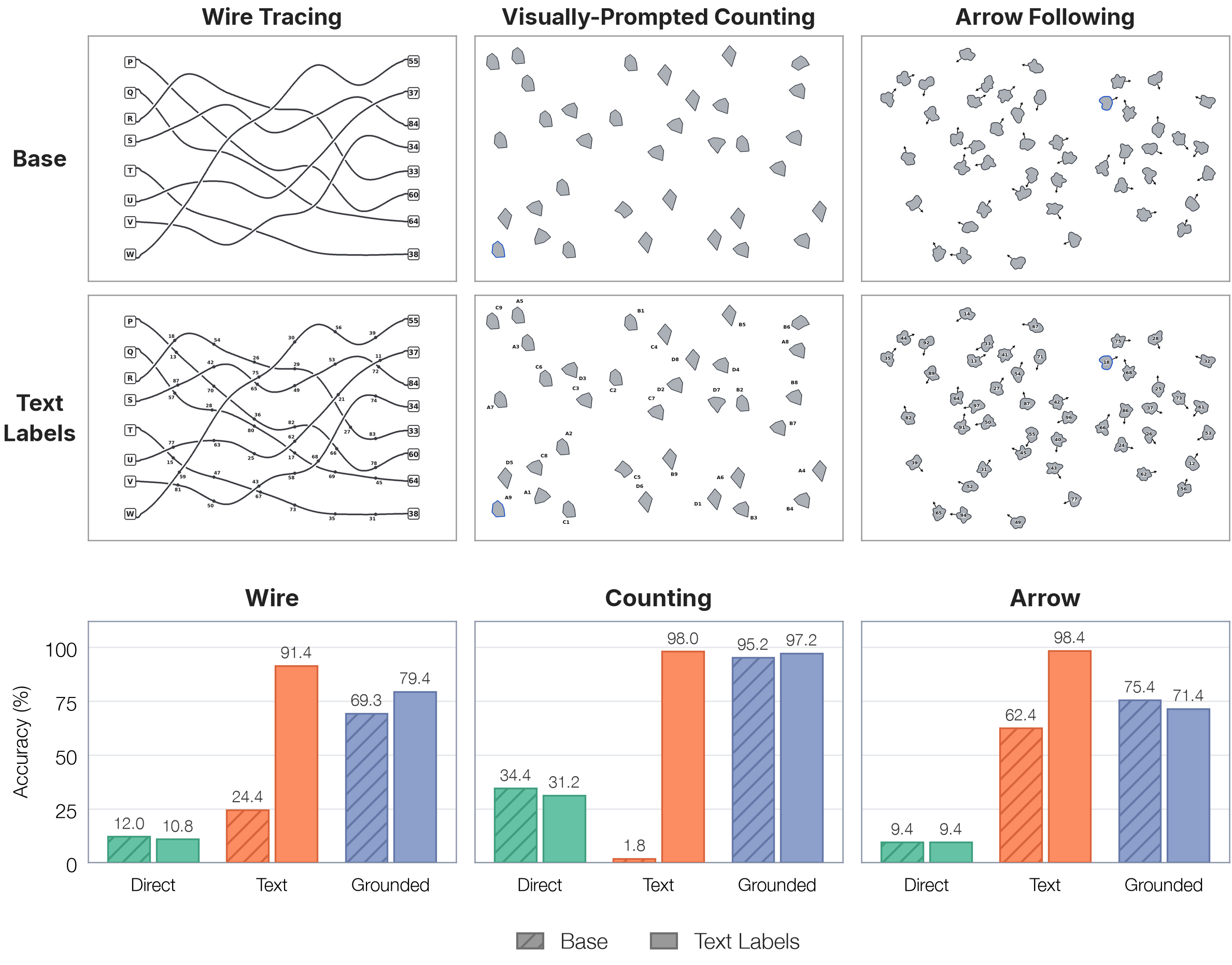}
    \caption{\textbf{Synthetic Probe Tasks}. We use three synthetic tasks to study where grounded reasoning improves over pure text reasoning. We find that grounding performs significantly better than text reasoning when no text-labels are available, but worse when given text labels. Zoom in for detail.}
    \label{fig:probe}
\end{figure}
\label{sec:study}
We hypothesize that grounding is primarily beneficial when the task requires an \textit{undescribable reference}. To illustrate this claim, we develop several synthetic tasks to probe where grounded reasoning improves over pure text reasoning. These tasks are designed to represent situations that are particularly difficult to solve with textual references. In particular, we use arbitrary shapes and a uniform color to prevent textual identification by common shape or color names. We also non-uniformly scatter objects, which makes maintaining the position of objects difficult using plain text.

\textbf{Probe tasks}. We consider three tasks, shown in Figure~\ref{fig:probe}. In \textit{wire-tracing}, the model is given a left terminal and must determine which right terminal its wire reaches. In \textit{visually prompted counting}, one polygon is outlined, and the model must count how many shapes are identical to it. In \textit{arrow following}, the model starts at an outlined blob and follows the arrows for several hops, before pointing to the final blob. Each task requires tracking undescribable visual references, such as the current position along a wire, the set of shapes already counted, or the current blob in the chain. Generation parameters and answer formats are given in Appendix~\ref{app:probe}.

\textbf{Experimental setup.} For each task, we finetune  Qwen3-VL-8B on 3 different reasoning variations: direct, text CoT, and grounded CoT. In the ``direct'' setting, no chain-of-thought is provided, and the model directly predicts the correct answer. In ``text CoT,'' we use plain text chain-of-thought, where positions of objects are described using language. In ``grounded CoT,'' we represent the positions of objects using [x, y] coordinates instead. To examine the benefit of grounding when positions in the image can be localized with language, we also re-render each task with text labels on objects or positions relevant for solving the task, as demonstrated in Figure \ref{fig:probe}. For these re-renders, we adjust the ``text CoT'' traces to refer to the precise text labels. Each variant is trained once and greedily decoded during evaluation.

\textbf{Grounded reasoning is helpful when text references are unavailable.} We report our results in Figure~\ref{fig:probe}. On the unlabeled scenes, we see that grounded reasoning improves significantly over text reasoning (69.3 vs 24.4 on wire-tracing) and direct fine-tuning (95.2 vs 34.4 on counting). However, once labels are added, the value of grounding over text reasoning drops significantly. We observe that text reasoning outperforms grounded reasoning on all three tasks when given text labels. As such, we conclude that grounding is primarily helpful when no textual references are available to solve a task.

\section{BrickVision}
\begin{figure}[h!]
    \centering
    \includegraphics[width=1\linewidth]{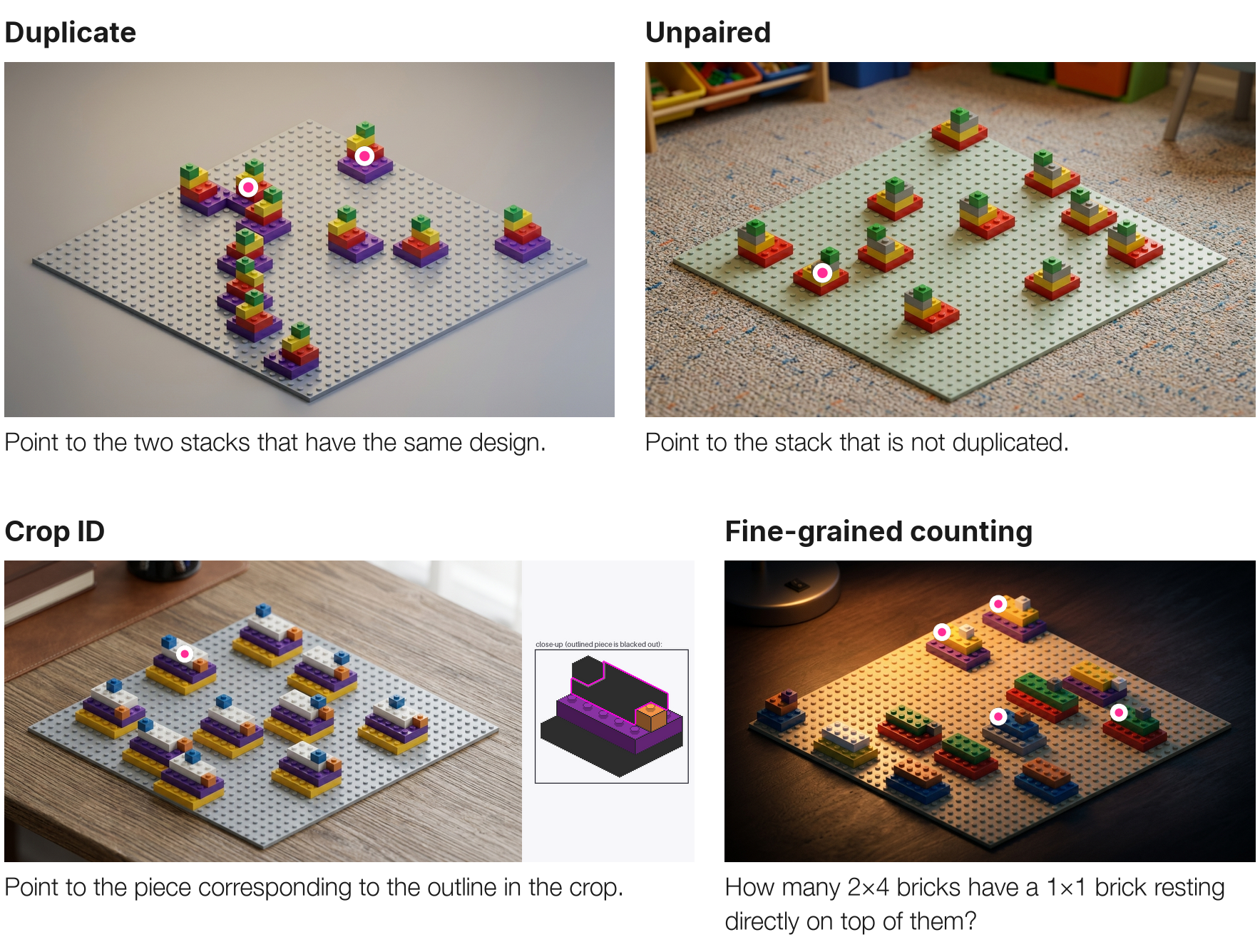}
    \caption{\textbf{BrickVision}. Samples from each task distribution of our BrickVision evaluation set. BrickVision tasks involve iterative comparison across objects difficult to describe in language, and require precise answers in the form of a point or integer.}
    \label{fig:BrickVision}
\end{figure}

Our study in Section \ref{sec:study} demonstrates that grounding capabilities are best illustrated on benchmarks that require undescribable references. However, because existing benchmarks have objects that are uniquely describable in text, they are easy to solve using textual references. To address this gap, we use our insights from Section~\ref{sec:why} to create BrickVision, a benchmark that requires undescribable references to solve.

\textbf{Benchmark construction}. BrickVision contains four different task types, described in Figure~\ref{fig:BrickVision}. Each task type has 250 samples, for a total of 1000 questions in the benchmark. Each task is designed to require iterative comparison over every pile of bricks. Both the precise geometry and position of the piles are difficult to describe in language, which necessitates the effective use of undescribable visual references.

To create BrickVision, we combine procedural generation with generative style transfer. For each task type, we create a procedural generator of isometric 2D scenes. We then pass this synthetic render into Nano Banana \citep{google2026nanobanana2} to restyle the image and use the original segmentation map as ground truth for the tasks that require point output (Appendix~\ref{fig:construction}). In order to use the original segmentation as ground-truth, the restyled scene must be aligned with the original synthetic render. However, we find that passing just the synthetic render to Nano-Banana results in misalignment of the segmentation masks, as shown in Appendix~\ref{fig:drift}. As such, we prompt the model with an additional render featuring bounding boxes of the block stacks and fix the aspect ratio during generation.

\begin{table}[h!]
\centering
\resizebox{\linewidth}{!}{%
\begin{tabular}{lccccc}
\toprule
Model & Crop ID & Fine-grained counting & Duplicate & Unpaired & Overall \\
\midrule
GPT-6 Astra \citep{openai2026gpt6astra}            & 90.0 & 25.2 & 74.8 & 84.8 & 68.7 \\ 
Claude Opus 5.5 \citep{anthropic2026opus55}        & 40.4 & 16.8 & 19.6 & 27.6 & 26.1 \\ 
GPT-5.6 Sol \citep{openai2026gpt56systemcard}      & 35.2 & 8.4  & 11.6 & 20.4 & 18.9 \\
Gemini 3.8 Flash \citep{google2026gemini38flash}   & 20.4 & 15.2 & 8.8  & 17.6 & 15.5 \\
Claude Opus 5 \citep{anthropic2026opus5systemcard} & 26.8 & 10.4 & 8.4  & 11.6 & 14.3 \\
Qwen3.8-Max \citep{qwen38}                        & 6.4  & 2.0  & 13.2 & 12.4 & 8.5  \\
\midrule
Qwen3.5-9B (base) \citep{qwen3.5} & 1.5 & 8.5 & 2.4 & 8.4 & 5.2 \\
\quad Finetuned Direct             & 8.1  & 16.0 & 8.3  & 7.7  & 10.0 \\
\quad Finetuned Text Reasoning     & 85.2 & 66.4 & 39.1 & 54.1 & 61.2 \\
\quad Finetuned Grounded Reasoning & 91.6 & 95.3 & 64.7 & 79.9 & 82.9 \\
\bottomrule
\end{tabular}%
}
\caption{\textbf{VLM performance on BrickVision}. Many frontier models struggle on BrickVision, suggesting limited ability to maintain undescribable visual references. Models trained with in-domain grounding data perform better than text reasoning, demonstrating the benefit of grounded reasoning.}
\label{tab:lego-base}
\end{table}

\textbf{BrickVision Evaluations.} We evaluate a variety of models on BrickVision and report our results in Table~\ref{tab:lego-base}. We find that many existing models struggle on BrickVision, with strong multimodal models such as Gemini 3.8 Flash scoring just 15.5\%. On the other hand, GPT-6 Astra is significantly stronger than all other models, scoring 68.7\% overall. We observe that Qwen 3.8 Max often points to empty regions of the image, unlike other frontier models which reliably point to a valid pile, which may suggest weaker overall grounding capabilities. More broadly, these results suggest that many existing frontier models have limited capability to reason over undescribable references. The performance disparity between GPT-6 Astra and other frontier models is particularly stark, and further work could attempt to analyze why this is the case.

Finally, to validate that the benchmark benefits from grounded reasoning, we fine-tune Qwen3.5-9B \citep{qwen3.5} on the pre-style-transfer images and ablate grounded and pure text reasoning strategies. We find that grounded reasoning significantly outperforms text-only reasoning (82.9 vs 61.2). Note that these fine-tuned results are intended as a diagnostic for the usefulness of grounding, as BrickVision is designed to evaluate models that have not been trained directly for the task.

\section{Conclusion} We have provided both a detailed analysis of \textit{where and why visual grounding is most useful} and a benchmark to help guide future research on grounded reasoning. However, we note an interesting conundrum: grounded reasoning is most useful on tasks that are difficult to describe in language, but automatic methods of acquiring grounding rely on text referring expressions. So how can we acquire labels for grounded reasoning at scale? We hope future work will shed light on this question.

\clearpage

\bibliography{references}
\bibliographystyle{iclr2026_conference}

\clearpage
\appendix

\section{Evaluations} 
\label{app:stats}
\textbf{Confidence Interval Calculations}. Following~\cite{miller2024addingerrorbarsevals}, we compute the question-level uncertainty (denoted $\text{SE}_{\text{q}}$) using a paired
standard error over per-question differences, averaging over checkpoints and clustering by image. We compute run-level uncertainty (denoted $s_{\text{run}}$) by taking the standard deviation of the delta over the three seeds. The final confidence interval is computed as
$\bar{\Delta} \pm 1.96\sqrt{\mathrm{SE}_{\text{q}}^{2} + s_{\text{run}}^{2}/3}$.

\section{Training Details}
\label{app:training}

Table~\ref{tab:hparams} lists the hyperparameters for the SFT and reinforcement learning stages. All runs are initialized from Qwen3-VL-8B Instruct unless otherwise noted. Following \cite{wang2026traceableevidenceenhancedvisual}, we use the AdamW \citep{loshchilov2019decoupledweightdecayregularization} optimizer for all stages and GRPO \citep{shao2024deepseekmathpushinglimitsmathematical} as the RL algorithm.

\begin{table}[h]
\centering
\footnotesize
\setlength{\tabcolsep}{3pt}
\begin{tabular}{lll}
\toprule
 & SFT & RL \\
\midrule
Epochs  & 1              & 5 \\
Batch         & 64             & 512 $\times$ 8 \\
LR            & $10^{-5}$      & $10^{-5}$ \\
Warmup        & cosine, 3\%    & constant \\
Weight decay  & 0              & 0.01 \\
KL / clip     & --             & 0.01 / 0.2 \\
\bottomrule
\end{tabular}
\caption{\textbf{Training hyperparameters}.}
\label{tab:hparams}
\end{table}

Our grounding reward is defined as follows:
{\setlength{\abovedisplayskip}{3pt}\setlength{\belowdisplayskip}{3pt}%
\begin{equation}
  r_j = \max_{i} s(p_i, b_j), \quad
  \mathrm{Rec} = \frac{1}{M}\sum_{j=1}^{M} r_j, \quad
  \mathrm{Prec} = \frac{1}{N}\sum_{j=1}^{M} r_j, \quad
  R_{\text{gnd}} = \frac{2\,\mathrm{Prec}\cdot\mathrm{Rec}}{\mathrm{Prec}+\mathrm{Rec}},
  \label{eq:rgnd}
\end{equation}}
where $r_j$ is the credit for the $j$-th annotated box,  $M$ is the number of annotated boxes $b_j$, $N$ is the number of distinct coordinates
$p_i$ emitted in the response. Because the ground truth is only annotated with boxes, instead of segmentation, we use a binary point-in-box credit during RL on pointing experiments. We use a Qwen3.5-9B judge model to judge the correctness of open-ended answers.

\section{Grounded SFT Data}
\label{app:sft-data}

\begin{figure}[h]
    \centering
    \includegraphics[width=1\linewidth]{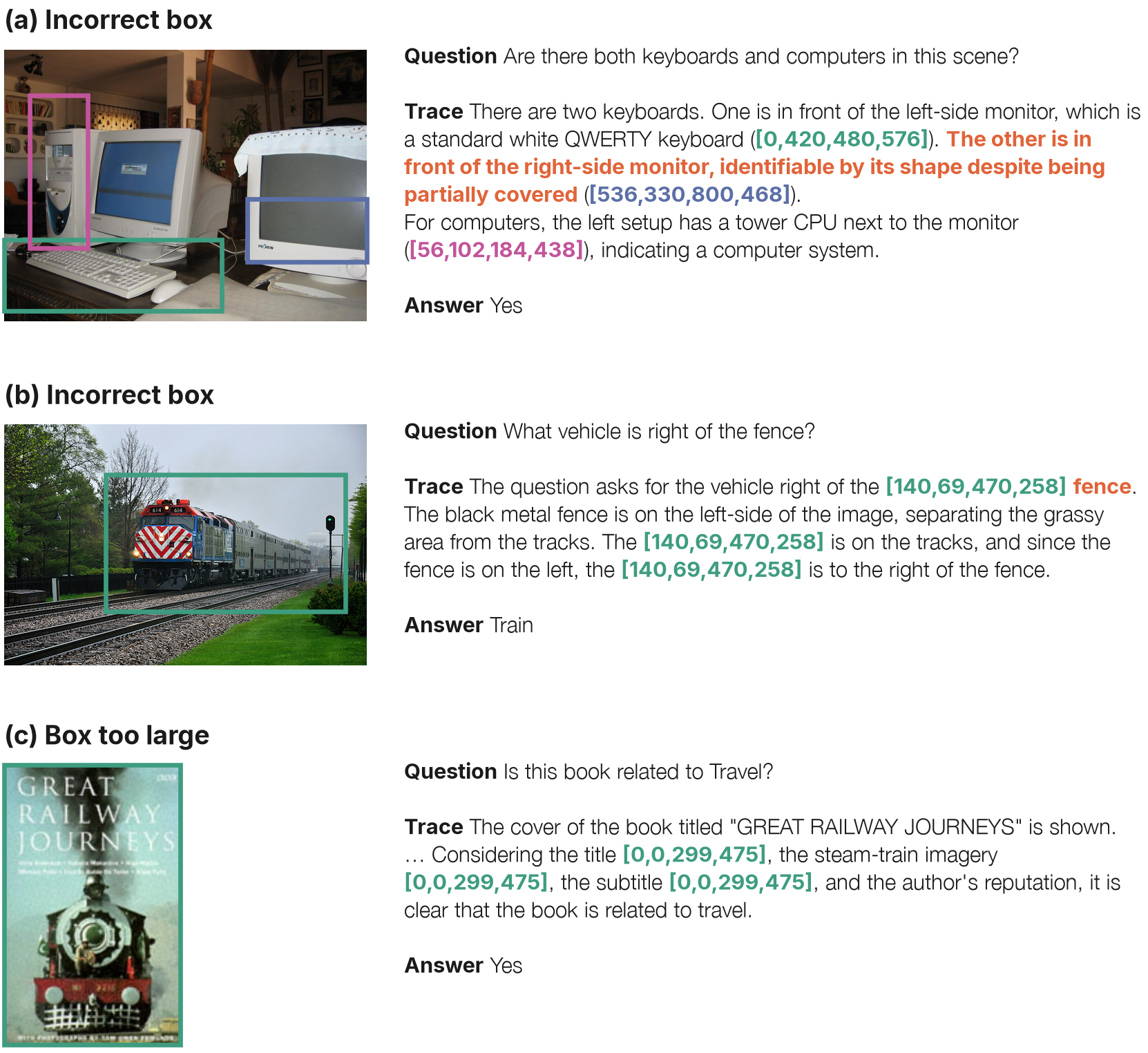}
    \caption{\textbf{Incorrect TreeVGR samples}. Selected samples from the TreeVGR dataset illustrating errors in data such as incorrect box labeling or uninformative boxes.}
    \label{fig:vgrsample}
\end{figure}

\textbf{Quality of existing data.} Figure~\ref{fig:vgrsample} shows TreeVGR SFT samples with boxes that are oversized or cover the wrong objects. As such, we regenerate our own data from ground truth derived from TVG \citep{zhang2026thinkingvisualgrounding}.

\textbf{Points versus boxes.} Table~\ref{tab:sft-pointbox} compares SFT checkpoints trained on point and box grounding with otherwise identical traces. Boxes performed significantly worse on SMQA, MHS, and MME-RW, so we use points in our SFT data. However, boxes are slightly better than points on counting.

\begin{table*}[h]
\centering
\begin{tabular*}{\textwidth}{@{\extracolsep{\fill}}lc cc c}
\toprule
Bench & $n$ & Point & Box & Box $-$ Point \\
\midrule
\multicolumn{5}{@{}l}{\emph{Spatial Reasoning}} \\
SMQA & 1076 & 0.527 & 0.509 & $\mathbf{-1.9 \pm 1.5}$ \\
MHS & 1500 & 0.539 & 0.525 & $\mathbf{-1.4 \pm 1.2}$ \\
EmbSpatial & 3640 & 0.733 & 0.728 & $-0.4 \pm 1.2$ \\
VSR & 1222 & 0.801 & 0.797 & $-0.4 \pm 2.1$ \\
OmniSpatial & 1533 & 0.470 & 0.467 & $-0.3 \pm 1.7$ \\
ERQA & 400 & 0.396 & 0.395 & $-0.1 \pm 2.1$ \\
\midrule
\multicolumn{5}{@{}l}{\emph{Fine-grained Perception}} \\
V$^*$ & 191 & 0.844 & 0.848 & $+0.3 \pm 2.3$ \\
MME-RW & 1919 & 0.531 & 0.521 & $\mathbf{-0.9 \pm 0.9}$ \\
TreeBench & 405 & 0.467 & 0.449 & $-1.8 \pm 2.4$ \\
\midrule
\multicolumn{5}{@{}l}{\emph{Counting}} \\
CountQA & 1528 & 0.327 & 0.351 & $\mathbf{+2.3 \pm 1.7}$ \\
FSC-147 & 1190 & 0.103 & 0.115 & $+1.2 \pm 2.0$ \\
\bottomrule
\end{tabular*}
\caption{\textbf{Point versus box grounding after SFT.} We report the box minus point accuracy delta with 95\% confidence intervals computed as in Appendix~\ref{app:stats}. Bold deltas exclude 0.}
\label{tab:sft-pointbox}
\end{table*}

\begin{figure}
    \centering
    \includegraphics[width=1\linewidth]{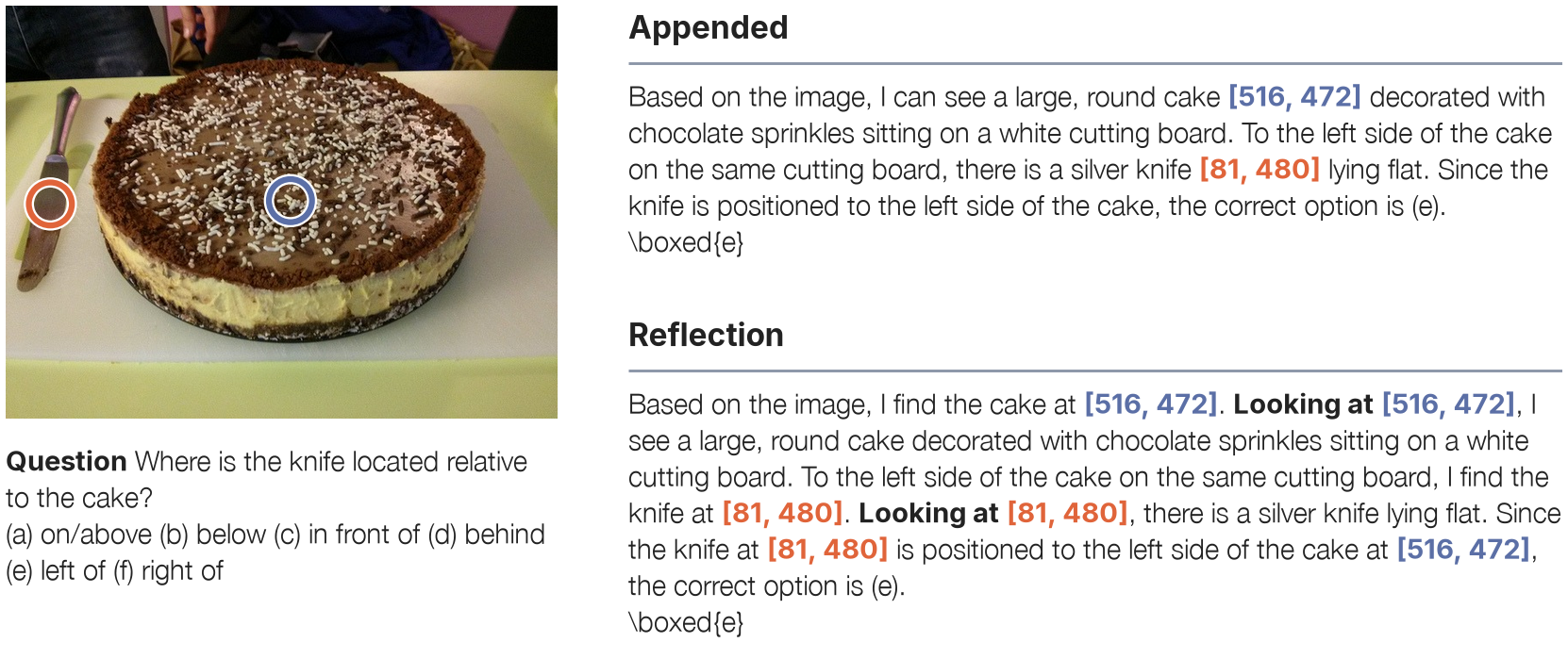}
    \caption{\textbf{SFT format ablation}. We compare the appended and reflection SFT formats. We find that the reflection format generally leads to better grounding usage, and use this format in our SFT data.}
    \label{fig:format}
\end{figure}

\textbf{Trace format.} Figure~\ref{fig:format} compares the Appended and Reflection trace formats. In Appended, coordinates are directly inserted into existing traces, while Reflection explicitly relooks at the coordinate.

\section{Probe Tasks}
\label{app:probe}

\textbf{Training details.} For wire-tracing and visually prompted counting, we finetune on training sets of 5k QA samples. Empirically, we found arrow-following to be a more difficult task and instead generate 50k QA samples for training. In each experiment, we train for 1 epoch. 

\textbf{Task generation.} We describe the generation procedure and answer format for each probe task.

\textit{Wire-tracing.} Each image has eight labeled terminals on both the left and right sides. Wires between terminals are generated with Catmull-Rom splines \citep{Catmull1974ACO}. Given the letter of a left terminal, the model must identify the right terminal that its wire reaches.

\textit{Visually prompted counting.} Each image contains $K \in \{29, 35, 41\}$ gray polygons. One polygon is outlined in blue and duplicated $M \in \{9, 10, 11, 12\}$ times. The model must output the number of shapes identical to the outlined one as an integer.

\textit{Arrow following.} Each scene contains 48 blobs, each with one arrow pointing to another blob. The start blob is outlined in blue. The model must follow the arrows for $H \in \{2, 4, 6\}$ hops and point to the final blob, answering with an [x, y] coordinate normalized to [0, 1000] following standard practice \citep{clark2026molmo2}.

\begin{figure}[h]
    \centering
    \includegraphics[width=1\linewidth]{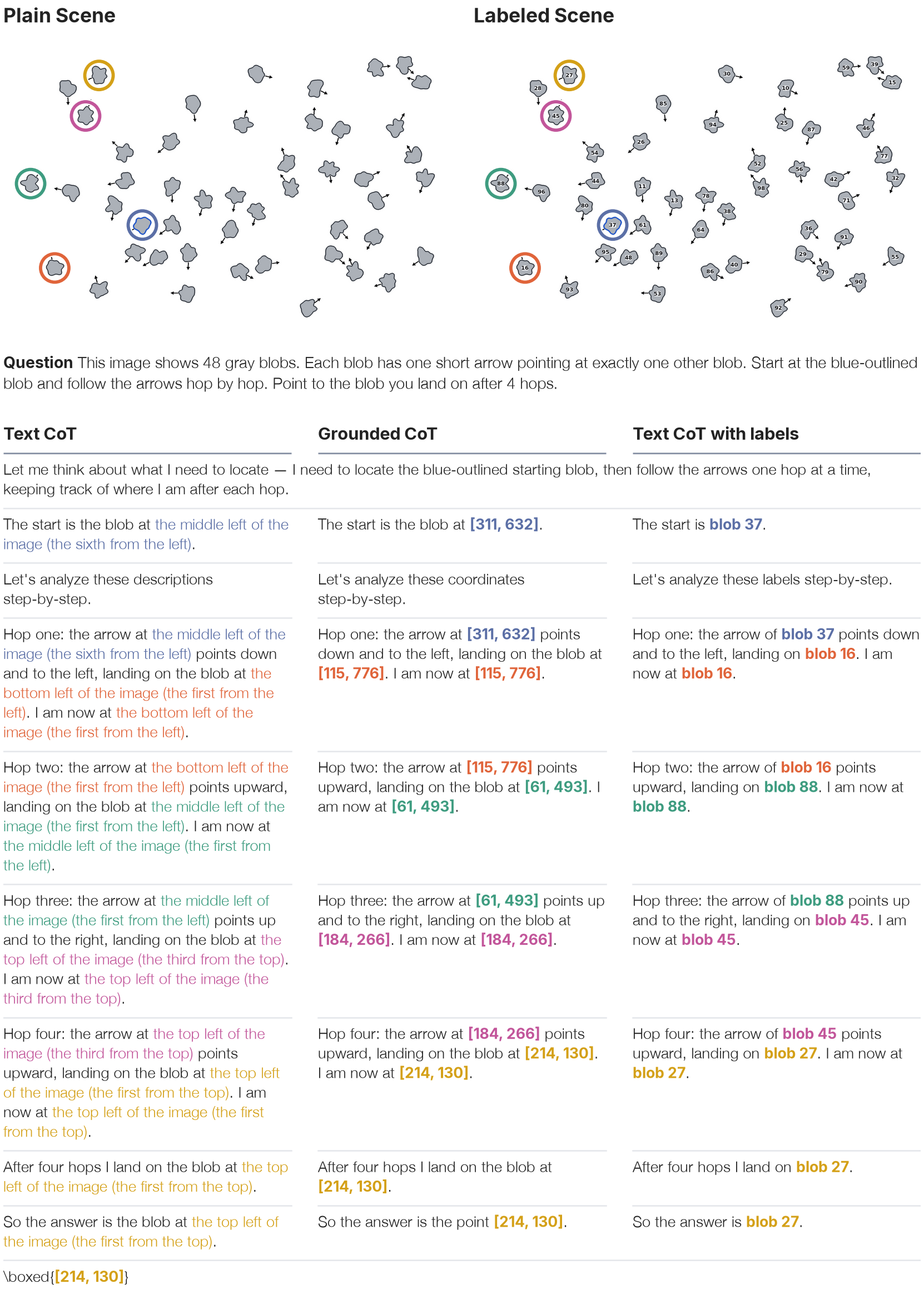}
    \caption{\textbf{Probe reasoning traces}. We provide examples of our reasoning data used in our ``Arrow Following'' task to illustrate the difference between  text traces and grounded traces.}
    \label{fig:probesft}
\end{figure}

\textbf{Chain-of-thought construction}. Figure~\ref{fig:probesft} shows paired text and grounded reasoning traces for arrow following. We use a similar strategy to create reasoning traces for the other probe tasks.

\section{BrickVision}
\label{app:BrickVision}
We consider four task types in BrickVision: Duplicate, Crop ID, Unpaired, and Fine-grained Counting.

\textbf{Duplicate.} Each scene has 10 to 12 stacks in the same colors, exactly two of which share a design, and the model must point to both. Every other stack differs from the shared design in the orientation or position of one or two bricks, so no single attribute separates the pair from the rest.

\textbf{Crop ID.} Each image shows a scene of 10 structures next to an enlarged inset of one of them. In the inset, two or three of the structure's five pieces are masked to silhouettes and one of the masked pieces is outlined. All structures use the same color for a given piece and differ only in the offset of a side marker and the position of the cap, so the model has to find the inset's structure in the scene by geometry and then point to the outlined piece.

\textbf{Unpaired.} Each scene has 10 to 13 stacks in the same colors. Every stack has an identical copy somewhere in the scene except one, and the groups of copies differ from each other only in geometry. The model must point to the stack that has no copy.

\textbf{Fine-grained Counting.} The task is to determine how many 2$\times$4 bricks satisfy a particular geometric predicate. We consider several predicates, such as having a 1$\times$1 brick on top, touching a 1$\times$1 at a short end, or resting crosswise on the brick below. The scenes also contain bricks of other shapes that satisfy the predicate, so the model must check each brick's shape as well as the predicate. Ground truth counts range from 2 to 7.

\begin{figure}[h]
    \centering
    \includegraphics[width=1\linewidth]{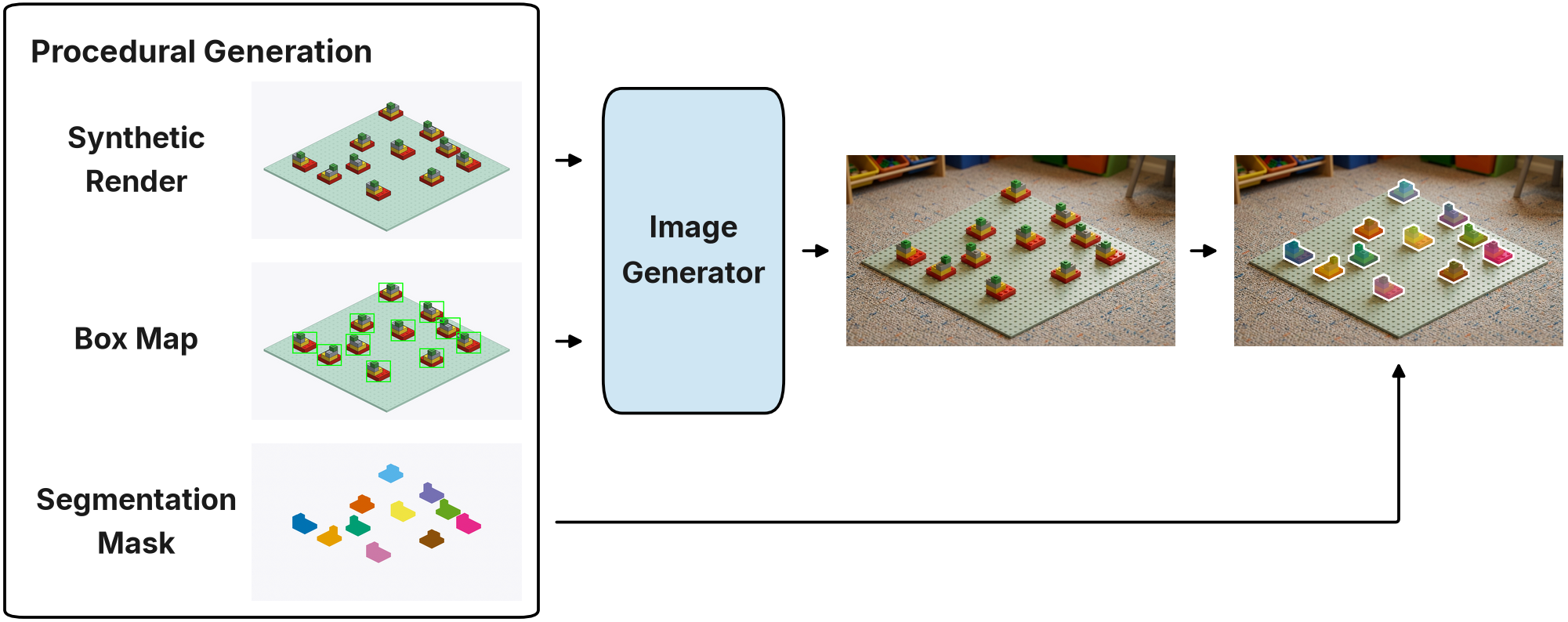}
    \caption{\textbf{Benchmark construction}. Our benchmark combines procedural generation with conditional image generation to produce realistic images aligned with the original synthetic ground truth.}
    \label{fig:construction}
\end{figure}

\begin{figure}[h]
    \centering
    \includegraphics[width=1\linewidth]{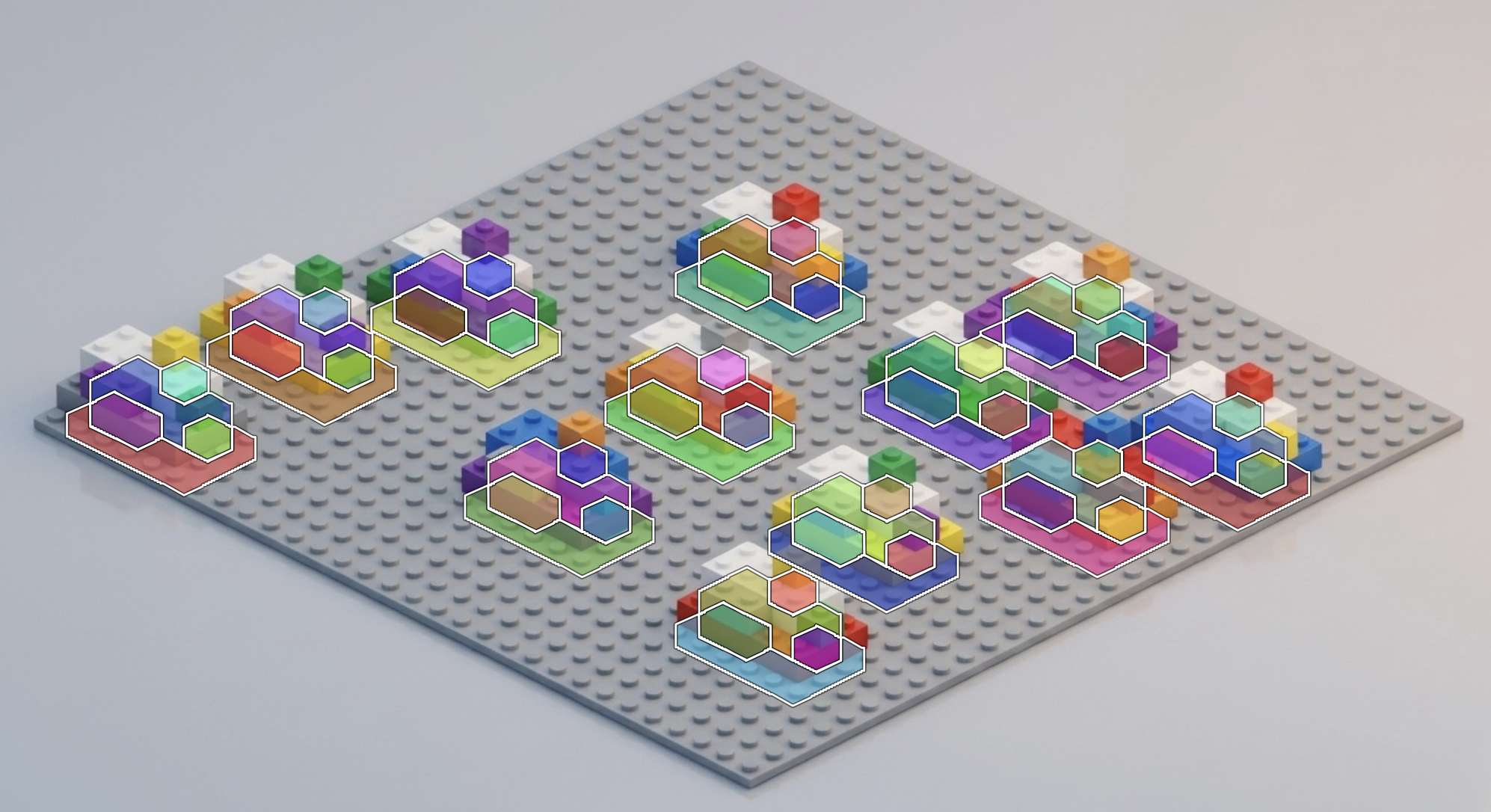}
    \caption{\textbf{Restyling drift}. Without locking the aspect ratio or prompting Nano-Banana with bounding boxes, we find that the restyled image drifts from the original, making the segmentation map ground truth (visualized as an overlay) incorrect.}
    \label{fig:drift}
\end{figure}

Figure~\ref{fig:construction} shows the pipeline used to generate the final benchmark images. A procedural generator renders an isometric scene and its segmentation map, and Nano-Banana restyles the render. Figure~\ref{fig:drift} shows the drift that occurs without a fixed aspect ratio and bounding-box conditioning.

\end{document}